\documentclass[letterpaper]{article} 
\usepackage{aaai24}  
\usepackage{times}  
\usepackage{helvet}  
\usepackage{courier}  
\usepackage[hyphens]{url}  
\usepackage{graphicx} 
\usepackage{natbib}  
\usepackage{caption} 
\usepackage{algorithm}
\usepackage{algorithmic}
\usepackage{graphicx}
\usepackage{amsmath}
\usepackage{amssymb}
\usepackage{booktabs}
\usepackage{multirow}
\usepackage{makecell}
\usepackage{multirow}
\usepackage{amsmath,amssymb}
\usepackage{adjustbox}
\usepackage{newfloat}
\usepackage{listings}
\DeclareCaptionStyle{ruled}{labelfont=normalfont,labelsep=colon,strut=off} 
\floatstyle{ruled}
\newfloat{listing}{tb}{lst}{}
\floatname{listing}{Listing}
\title{FuRPE: Learning Full-body Reconstruction from Part Experts}
\author{
     Zhaoxin Fan\textsuperscript{\rm 1},
     Yuqing Pan\textsuperscript{\rm 1},
     Hao Xu\textsuperscript{\rm 2},
    Zhenbo Song\textsuperscript{\rm 3}\\
    Zhicheng Wang\textsuperscript{\rm 4},
    Kejian Wu\textsuperscript{\rm 4},
    Hongyan Liu\textsuperscript{\rm 5},
    Jun He\textsuperscript{\rm 1}\thanks{Corresponding author: hejun@ruc.edu.cn}\\
    \textsuperscript{\rm 1}Renmin Universtiy of China, 
    \textsuperscript{\rm 2}Psyche AI Inc,\\
    \textsuperscript{\rm 3}Nanjing University of Science and Technology,
    \textsuperscript{\rm 4}Xreal,\\
    \textsuperscript{\rm 5}Tsinghua University
    \\
}

\begin{document}

\maketitle

\begin{abstract}
In the field of full-body reconstruction, the scarcity of annotated data often impedes the efficacy of prevailing methods. To address this issue, we introduce FuRPE, a novel framework that employs part-experts and an ingenious pseudo ground-truth selection scheme to derive high-quality pseudo labels. These labels, central to our approach, equip our network with the capability to efficiently learn from the available data. Integral to FuRPE is a unique exponential moving average training strategy and expert-derived feature distillation strategy. These novel elements of FuRPE not only serve to further refine the model but also to reduce potential biases that may arise from inaccuracies in pseudo labels, thereby optimizing the network's training process and enhancing the robustness of the model. We apply FuRPE to train both two-stage and fully convolutional single-stage full-body reconstruction networks. Our exhaustive experiments on numerous benchmark datasets illustrate a substantial performance boost over existing methods, underscoring FuRPE's potential to reshape the state-of-the-art in full-body reconstruction.
\end{abstract}

\begin{figure}[h]
  \centering
      \includegraphics[width=0.85\linewidth]{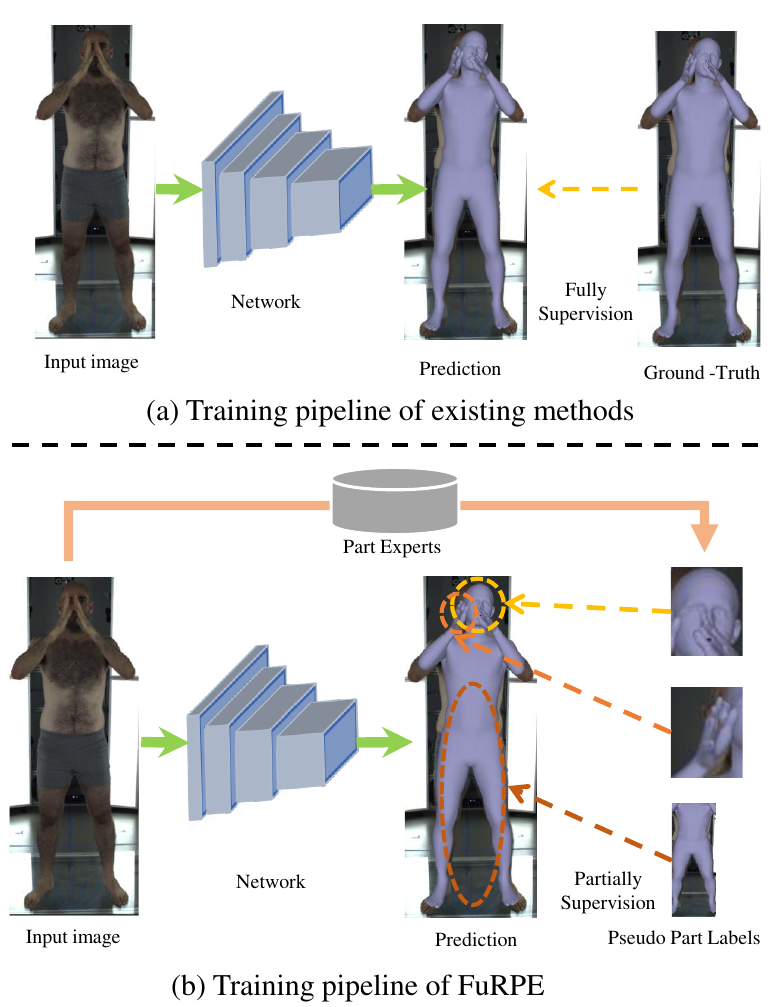}
 \vspace{-0.1in}
  \caption{(a) Traditional methods train on costly, scarce annotated data. (b) Our method utilizes affordable, high-quality pseudo labels from part-experts.}
  \label{fig:motivation}
  \vspace{-0.28in}

\end{figure}

\section{Introduction}
\label{sec:intro}
Interpreting human behavior and appearance from real-world imagery and video data is a crucial prerequisite for a multitude of applications, with robotics \cite{dupont2021decade,anwar2019systematic,ortenzi2021object} and augmented reality \cite{fan2021deep,xiong2021augmented,siriwardhana2021survey} standing out as prime examples. The task of human body reconstruction, a critical yet challenging aspect within computer vision, substantially contributes to this understanding. It involves estimating a mesh from a given image or video that aptly represents the target human's pose and appearance, serving as a cornerstone in applications that hinge on digital human representations.

The landscape of human body reconstruction is enriched with diverse methodologies \cite{kolotouros2019learning,kocabas2020vibe,choi2021beyond,rong2022chasing,zhang2021pymaf}. These can be broadly bracketed into two categories: methods that predict the mesh topology directly from input images, and those that estimate the parameters of the parametric human body model SMPL \cite{loper2015smpl}. Despite significant strides, these methods have a common limitation - they primarily focus on reconstructing only the body parts, thereby constraining their broader applicability. To address this limitation, several recent works \cite{choutas2020monocular,rong2021frankmocap,moon2020pose2pose,zhou2021monocular,lin2023one} have embarked on full-body reconstruction methods. These methodologies harness the power of the SMPL-X model \cite{pavlakos2019expressive}, which incorporates the head and hands into SMPL's representation. However, these methods encounter a significant challenge: the scarcity of adequately annotated data for training.  This challenge originates from the intricacies associated with annotating full-body parameters from an image. Conventional methods involve using optical or inertial tools for recording the body's pose. However, capturing the head and hands requires advanced, more expensive equipment like the Vrtrix, adding a layer of financial burden and complexity to the process. Furthermore, harmonizing the data from these three distinct captures into a cohesive representation is a significant challenge.

In this paper, we present FuRPE, a novel method designed to tackle the complexities of annotating full-body parameters from an image. FuRPE harnesses the potential of the SMPL-X model, which can be decomposed into three sub-parametric models: SMPL \cite{loper2015smpl}, FLAME \cite{li2017learning}, and MANO \cite{romero2022embodied}. Several part experts \cite{kolotouros2019learning,kocabas2020vibe,choi2021beyond,ge20193d,zhou2020monocular,feng2021learning,danvevcek2022emoca} have shown promising results for these individual components. As illustrated in Fig. \ref{fig:motivation}, adopting the concept from \cite{weinzaepfel2020dope}, FuRPE utilizes these experts to generate pseudo full-body ground-truth labels, enabling us to exploit large-scale datasets for training a full-body reconstruction model. While \cite{weinzaepfel2020dope} predicts only the 3D keypoints of the body, FuRPE extends this by predicting SMPL-X parameters with improved pseudo labels and carefully crafted modules. Although this makes our task more challenging, it is also more practical, considering the full-body shape and appearance, and enabling a broader range of applications.

Specifically, FuRPE ingeniously leverages the pseudo labels produced by part experts. This approach significantly augments the valuable training data available, leading to an enhancement in the overall performance of the system. To ensure the quality of these pseudo labels, a simple yet highly effective pseudo ground-truth selection scheme is employed. This scheme plays a pivotal role in refining the training process and mitigating potential bias that could be introduced by inaccurate pseudo labels. Alongside the pseudo ground-truth selection scheme, FuRPE adopts an Exponential Moving Average (EMA) training strategy \cite{klinker2011exponential} and an expert-derived feature distillation strategy. These strategies are designed to refine the training outcomes and further curtail the bias induced by inaccurate pseudo labels. The EMA training strategy is especially noteworthy as it promotes a self-supervised joint training process between student-teacher networks, which significantly bolsters the training performance. Crucially, at the inference stage, FuRPE only requires the student model to predict the full-body SMPL-X parameter.

We apply FuRPE to train both two-stage and fully convolutional single-stage full-body reconstruction networks, using a variety of well-established human body reconstruction datasets. This application showcases the method's adaptability and robustness across different network structures. The empirical results from these comprehensive experiments significantly underscore FuRPE's superiority over the baseline models, thereby establishing a new benchmark in the field of full-body reconstruction performance. A consistent improvement in the performance of our model was observed as the size of the training dataset increased. This trend effectively demonstrates FuRPE's capability to utilize existing publicly available datasets in a highly efficient manner. 

Our contributions can be summarized as follows: 1) We present FuRPE, an innovative method the full-body reconstruction task, which notably amplifies the performance of both two-stage and one-stage strong baselines. 2) We propose to harness knowledge from part-experts for full-body reconstruction, complemented by a simple yet effective pseudo ground-truth selection scheme. In addition, we incorporate an Exponential Moving Average training strategy and introduce an expert-derived feature distillation strategy to reduce bias exist in pseudo labels. 3) We conduct extensive experiments on several publicly available datasets, demonstrating the effectiveness of using pseudo labels and  expert-derived features for full-body reconstruction.

\begin{figure*}[t]
  \centering
  \includegraphics[width=0.9\linewidth]{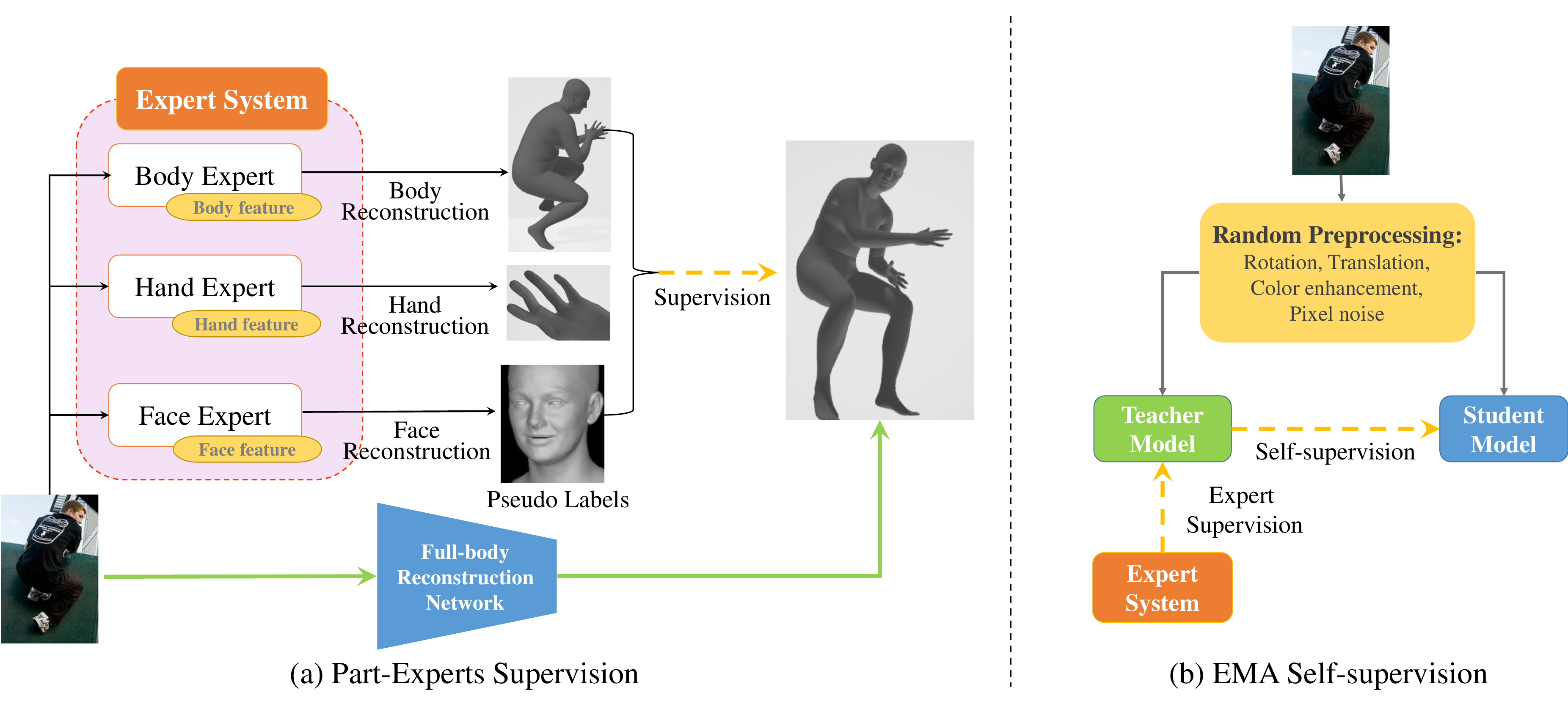}
 \vspace{-0.1in}
  \caption{Pipeline of our work. (a) The training pipeline of using part-experts to generate supervision signals. (b) The training pipeline of Exponential Moving Average self-supervision.}
  \label{fig:pipeline}
   \vspace{-0.2in}
\end{figure*}
\section{Related Works}
\subsection{3D Human Pose Estimation}
3D human pose estimation is closely related to body mesh reconstruction, as both tasks involve predicting the 3D keypoints of a person from an image. There are two main categories of 3D human pose estimation methods: direct prediction methods and 2D-3D lifting methods. Direct prediction methods \cite{moon2019camera,mehta2018single,pavlakos2017coarse,rogez2019lcr} directly predict the 3D keypoints from the input image. For example, LCR-Net++ \cite{rogez2019lcr} predicts 2D and 3D poses of multiple people simultaneously to increase the robustness of 3D keypoints estimation from a fully-connected prediction branch. On the other hand, 2D-3D lifting methods \cite{martinez2017simple,pavllo20193d,zhan2022ray3d,li2022exploiting} first estimate the 2D keypoints of the human and then use a lifting network to lift these keypoints to 3D keypoints. For instance, \cite{li2022mhformer} proposes to predict several hypotheses from 2D keypoints for 3D keypoints prediction.
Despite significant progress, 3D human pose estimation methods only estimate the keypoints of a human and neglect the prediction of shape and appearance. In contrast, our proposed method reconstructs the parameters of the SMPL-X model, which represent a human's pose and shape parameters. This allows for a more comprehensive representation of the human body, enabling a wider range of applications.

\subsection{3D Human Body Reconstruction}
3D human body reconstruction aims to predict the mesh of a target human from a single image or video. HMR \cite{kanazawa2018end} recovers SMPL parameters from a single image using a model trained on pseudo labels generated from SMPLify \cite{loper2015smpl}. SPIN \cite{kolotouros2019learning} proposes to reconstruct 3D human pose and shape via model-fitting in the loop, making the training pipeline self-supervised. Both TCMR \cite{choi2021beyond} and VIBE \cite{kocabas2020vibe} introduce the use of temporal information for 3D human body reconstruction to improve stability and smoothness of the prediction results. Many subsequent works \cite{lin2021end,rempe2021humor,kocabas2021pare} have been proposed to improve the performance of SMPL parameter estimation. Some methods \cite{alldieck2021imghum} also estimate the parameters of other parametric models such as GHUM \cite{xu2020ghum}. However, these methods only predict the mesh of body parts, limiting their broader application. In contrast, our paper focuses on the more general task of full-body reconstruction.

\subsection{Full-body Reconstruction}
Building upon SMPL-X's introduction \cite{pavlakos2019expressive}, several strategies offer full-body reconstruction. SMPLify-X, an optimization-based method, suffers from impractical speed. ExPose \cite{choutas2020monocular} utilizes a neural network for direct full-body SMPL-X parameters prediction from a single image and introduces a body-attention mechanism for interrelations among head, body, and hands. FrankMocap \cite{rong2021frankmocap} suggests a strategy that stitches together different body parts to form whole-body parameters. Concurrently, PIXIE \cite{feng2021collaborative} merges features from different part modules in the network through a moderator, allowing all parts to contribute to the whole-body reconstruction. Despite acceptable performance, these methods face limitations due to the scarcity of well-annotated training data. OSX \cite{lin2023one} introduces an optimization-based method for whole-body parameters annotation, but it remains computationally expensive. Contrarily, our method efficiently generates pseudo labels and features from part experts and leverages large-scale public datasets for training, significantly enhancing full-body reconstruction models' performance. Our work resembles \cite{weinzaepfel2020dope}, which also use pseudo labels generation to predict 3D human body parameters. However, their approach predicts only the full-body 3D keypoints, while ours estimates both human pose and shape and employs an Exponential Moving Average strategy for further performance improvement.

\section{Method}
\subsection{Overview}
In this section, we delve into the comprehensive description of our proposed methodology, FuRPE. The process flow of our method is illustrated in Fig. \ref{fig:pipeline}. As depicted in Fig. \ref{fig:pipeline} (a), the process commences with the generation of pseudo labels and pseudo features. We denote the pseudo labels and features for the face, hand, and body as $(L^f_{pre}, F^f)$, $(L^h_{pre}, F^h)$, and $(L^b_{pre}, F^b)$ respectively before the selection process.

To ensure high-quality training data, we implement an elaborate pseudo ground-truth selection scheme, which refines these initial pseudo labels to $(L^f, F^f)$, $(L^h, F^h)$, and $(L^b, F^b)$, where the subscript indicates the corresponding body part (face: 'f', hand: 'h', body: 'b').

As shown in Fig. \ref{fig:pipeline} (b), we introduce an Exponential Moving Average (EMA) self-supervision pipeline to further enhance performance. At the onset of training, we simultaneously instantiate a student network and a teacher network. We denote an input image as $I$, which is randomly augmented into two images, $I_a$ and $I_b$. $I_a$ is processed by the teacher network, and $I_b$ is processed by the student network. The outputs and weights of the two networks should align according to the data augmentation method. To maintain this consistency, we adopt an EMA training strategy to update the weights of the networks, enabling them to be trained jointly. During inference, only the student network is utilized to predict the full-body parameters.

\subsection{Part Experts and Pseudo labels}

The task of integrating part-specific annotations from diverse devices into a unified full-body annotation is addressed by leveraging the separability of the SMPL-X model. It is segregated into the FLAME model for the face, SMPL model for the body, and the MANO model for the hands. We utilize the predictions of part-expert models, trained on part-specific annotated data, as pseudo part labels. Accordingly, we introduce three part-experts in our work, each dedicated to a specific part: face, body, and hands. These experts - SPIN, DECA, and FrankMocap - are well-trained deep learning models that perform robust part-specific predictions from respective cropped images.

\noindent \textbf{Body Expert:} We utilize the SMPL model as our body expert. The pose parameters $\theta_{body} \in R^{72}$ and shape parameters $\beta_{body} \in R^{10}$ of the body are estimated from the cropped body image $I_{body}$ with the SPIN model, which has been well-trained on a large dataset comprising diverse body shapes and poses.

\begin{equation}
{\theta_{body}, \beta_{body}} = f_{SPIN}(I_{body})
\end{equation}

\noindent \textbf{Face Expert:} The FLAME model is employed for face reconstruction. The parameters for facial identity $\beta_{face} \in R^{200}$, pose $\theta_{face} \in R^{3k+3}$ (where $k=4$ joints: neck, jaw, and eyeballs), and expression $\psi_{face} \in R^{100}$, are estimated from the cropped face image $I_{face}$ with the DECA model, which has been well-trained on a large dataset comprising diverse facial expressions and identities.

\begin{equation}
{\beta_{face}, \theta_{face}, \psi_{face}} = f_{DECA}(I_{face})
\end{equation}

\noindent \textbf{Hand Expert:} The MANO model is used for hand reconstruction. The pose parameters $\theta_{hand} \in R^{21 \times 3}$ and shape parameters $\beta_{hand} \in R^{10}$ of the hands are estimated from the cropped hand image $I_{hand}$ with the FrankMocap model, which has been well-trained on a large dataset comprising diverse hand poses and shapes.

\begin{equation}
{\beta_{hand}, \theta_{hand}} = f_{FrankMocap}(I_{hand})
\end{equation}

\subsection{Pseudo Ground-truth Selection  Scheme}

Though the three part-experts can generate expressive pseudo labels and pseudo features, we can't guarantee that every pseudo ground-truth is in high quality. To filter our low quality pseudo ground-truths, we propose a three-step pseudo label select scheme.

\noindent \textbf{Step 1:} We leverage Openpose to detect the 2D key points of each person in the image. Then, we count the number of high confident keypoints. When the number of high confident keypoints is smaller than 12, we discard this image. The confidence thresholds are 0.1, 0.2, 0.4 for body, hand, and face respectively. This process can be expressed as below:

\begin{equation}
K_{2D} = \text{OpenPose}(I), \, n_{conf} = \text{count}(K_{2D}),
\end{equation}

\begin{equation}
\text{discard } I \text{ if } n_{conf} < 12.
\end{equation}

\noindent \textbf{Step 2:} The left images are input into our part-experts for parameters estimation. We discard these images with invalid output. This process can be expressed as below:

\begin{equation}
\theta, \beta = \text{PartExperts}(I),
\end{equation}

\begin{equation}
\text{discard } I \text{ if output is invalid.}
\end{equation}

\noindent \textbf{Step 3:} We use the predicted part parameters to drive the SMPL-X model and get the 3D keypoints. These 3D keypoints are projected into 2D keypoints. Then, we calculate the mean square error (MSE) between these projected 2D keypoints and 2D keypoints detected by OpenPose. If the MSE is larger then 1.5cm, we discard this image. This process can be expressed as below:

\begin{equation}
K_{3D} = \text{SMPLX}(\theta, \beta), \, K_{\text{proj}} = \text{Project}(K_{3D}),
\end{equation}

\begin{equation}
\text{MSE} = \text{mean}((K_{\text{proj}} - K_{2D})^2),
\end{equation}

\begin{equation}
\text{discard } I \text{ if MSE } > 1.5.
\end{equation}

The final left images and part-parameters are used for the full-body reconstruction model training.

\begin{table*}[ht]
\centering
\caption{Comparison on EFH dataset.}
 \vspace{-0.1in}
\scalebox{0.8}{
\begin{tabular}{c|c|c|c|c|c|c}
\hline
{\bf Category} & {\bf Methods} & {\bf V2V/Procrustes} & {\bf PA-V2V Body} & {\bf PA-V2V L-Hand} & {\bf PA-V2V R-Hand} & {\bf PA-V2V Face} \\
\hline
Baseline & Expose  & 55.1 & 52.9 & 13.1 & 12.6 & 5.7 \\
\hline
\multirow{4}{*}{SOTA Methods} & Frankmocap(copy \& paste) & 58.2 & 52.7 & 10.9 & 11.2 & 5.7 \\
 & Frankmocap(optimization) & 54.7 & 53.3 & {\bf 10.8} & 11.2 & 5.4 \\
 & Frankmocap(neural network) & 57.1 & 53.2 & 10.9 & 11.2 & 5.7 \\
 & PIXLE  & 55.0 & 53.0 & 11.0 & 11.2 &  {\bf 4.6} \\
\hline
\multirow{2}{*}{Ours} & FuPRE+Expose & {\bf 50.6} & {\bf 51.6} & 12.4 & 11.7 & 5.1 \\
 & FuPRE+ResNet & 54.8 & 52.9 & 12.4 & 12.1 & 5.6 \\
\hline
\end{tabular}
}
\label{tab:efh}
\end{table*}

\subsection{Training using Pseduo Labels}

To extract knowledge from the pseudo labels generated by part experts, we compute three distinct loss types: body loss, head loss, and hand loss to train a full-body reconstruction network. 

Body loss is composed of the Mean Absolute Error (MAE) between 2D body joints and the Mean Square Error (MSE) between body poses. The head and hand losses follow a similar structure, but an additional expression loss is included in the head loss. 

The overall training loss can be expressed as follows:

\begin{align*}
L_{\text{total}} &= L_{\text{body}} + L_{\text{face}} + L_{\text{hand}}, \\
L_{\text{body}} &= L_{\text{2d-body-joint}} + L_{\text{pose}}, \\
L_{\text{face}} &= L_{\text{2d-face-joint}} + L_{\text{expression}} + L_{\text{jaw-pose}}, \\
L_{\text{hand}} &= L_{\text{2d-hand-joint}} + L_{\text{hand-pose}}.
\end{align*}

In the above loss function, the 2D joint loss ($L_{\text{2d-body-joint}}$, $L_{\text{2d-face-joint}}$, and $L_{\text{2d-hand-joint}}$) is calculated as:

\begin{equation}
L_{\text{2d-joint}} = \sum_{j=1}^J v_j \left\| \hat{x_j} - x_j \right\|_1,
\end{equation}

where $v_j$ is the binary variable representing the visibility of the $j$th joint, $\hat{x_j}$ refers to the ground truth value, and $x_j$ is the predicted value.

The pose loss can be calculated as:

\begin{equation}
L_{\text{pose}} = \left\| \hat{\theta} - \theta \right\|_2^2,
\end{equation}

where $\hat{\theta}$ refers to the ground truth value, and $\theta$ is the predicted value.

The expression loss follows the same function, with $\phi$ representing ground truth expression parameters:

\begin{equation}
L_{\text{expression}} = \left\| \hat{\phi} - \phi \right\|_2^2.
\end{equation}

\subsection{Expert-Derived Feature Distillation Strategy}

Apart from the pseudo labels, we also employ pseudo features generated by part experts for training our full-body reconstruction model, as depicted in Fig. \ref{fig:pipeline}. This strategy is not because these expert-derived features are inherently superior, but because they can supplement the information that may be lost in the pseudo labels.

By distilling the knowledge embedded in these expert-derived features, we can effectively guide the training of our model. This approach refines the training outcomes and reduces the bias that inaccurate pseudo labels may induce.

The feature loss for each body part is computed separately. We denote the feature loss for the body part as $L_{b-feature}$. The feature losses for the head and hand parts follow a similar structure and are calculated as follows:

\begin{equation}
L_{feature} = A \cdot KL(\hat{f}, f),
\end{equation}

where $KL$ represents the Kullback-Leibler divergence, $A$ is an amplification coefficient empirically set as $e^5$, $\hat{f}$ is the pseudo feature generated by the part experts, and $f$ is the predicted feature of our model.


\subsection{Exponential Moving Average Training Strategy}

In our training framework, depicted in Fig. \ref{fig:pipeline} (b), we include an Exponential Moving Average (EMA) self-supervision strategy to further refine the training outcomes and diminish the bias brought by imprecise pseudo labels. This approach is inspired by Huang et al. \cite{huang2021spatio}.

We initialize two networks: a teacher network and a student network. The consistency between their outputs, under different random data augmentations, provides a self-supervision mechanism. The teacher model serves as the regression target for the student model, with their parameters updated by the training loss and an EMA process, respectively.

The parameters of the teacher network ($\sigma$) and the student network ($\phi$) are updated as follows:

\begin{equation}
\phi \rightarrow \tau\phi + (1-\tau)\sigma,
\end{equation}

where $\tau \in [0,1]$ is the decay rate of the moving average.

The EMA loss during training is computed as:

\begin{equation}
L_{EMA}= L_{total} + L_{o\rightarrow t} + L_{t\rightarrow o},
\end{equation}

where $L_{o\rightarrow t}$ denotes the student-to-teacher loss, and $L_{t\rightarrow o}$ is its inverse, computed as:

\begin{equation}
L_{a\rightarrow b}= 2-2\cdot{\frac{\langle z_a,z_b \rangle}{\left\|z_a\right\|_2 \cdot \left\|z_b\right\|_2}},
\end{equation}

with $z_a$ representing the parameters predicted by model $a$ (i.e., the student model).

\begin{table*}[t]
\centering
\caption{Performance Improvements through Incremental Data Addition.}
\vspace{-0.1in}
\scalebox{0.9}{
\begin{tabular}{c|ccccc}
\hline
   Methods & V2V/Procrustes & PA-V2V Body & PA-V2V Left Hand & PA-V2V Right Hand & PA-V2V Face \\
\hline
curated fittings &       53.3 &       52.4 &         13 &       12.4 &        5.3 \\

      +mpi &       52.4 &       51.7 &         13 &       12.7 &        5.3 \\

     +3DPW &       52.3 &       51.7 &         13 &       12.8 &        5.3 \\

+Human3.6m &       51.4 &       51.7 &       12.7 &       11.8 &        5.1 \\

  +ochuman &       51.1 &       51.6 &       12.6 &       11.8 &        5.1 \\

+posetrack  &       50.9 &       51.6 &       12.7 &       11.8 &        5.1 \\

      +EFT & {\bf 50.6} & {\bf 51.6} & {\bf 12.4} & {\bf 11.7} &  {\bf 5.1} \\
\hline
\end{tabular}  
\label{tab:data}
}
\end{table*}

\section{Experiment}

\subsection{Implementation Details}
We implement our method using Pytorch. The model is optimized using the Adam optimizer. We train two models in our implementation. The first one is our baseline method: the ExPose Network, which uses three sub networks for body, head and hand parameters prediction respectively. The second one is setted up by ourselves. In this model, only a sub network is used for all the body, head and hand reconstruction. Therefore, this model is more light weight than  the ExPose Network, termed as a one stage method. The initial  learning rate is  0.00001, then reduces by 10 times after training for 20 epochs. The batch size is set as 32.  All the experiments are conducted on a single A6000 GPU. 


\subsection{Dataset}
\noindent \textbf{Training Data}.The training datasets that we used include CuratedFits \cite{choutas2020monocular}, H3.6M\cite{Ionescu2014human}, MPI\cite{mehta2017monocular}, Ochuman \cite{zhang2019pose2seg}, Posetrack \cite{joo2021exemplar}, and EFT\cite{joo2021exemplar}. Among them, the first three ones are indoor datasets while the remaining are outdoor images, which consist of diversified pose sources and scenarios. After data selection, H3.6M contains 234041 pictures, MPI contains 133522 pictures, Ochuman contains 3936 pictures, Posetrack contains 22895 pictures, and EFT contains 2074 pictures.

\noindent \textbf{Evaluation Data}. In order to compare our model with the state-of-the-art 3D human pose estimation models especially targeting at head, hand and body, together with several recently proposed full-body reconstruction models, we use several widely used public available test sets for experiments. For part-only evaluation, NoW \cite{feng2018evaluation} is used for the head evaluation, FreiHAND \cite{zimmermann2019freihand} is used for  hand evaluation, and 3DPW \cite{martinez2017simple} is used for body evaluation. As for full-body evaluation, we use EHF \cite{pavlakos2019expressive} to test the comprehensive capture ability of the model and the effectiveness of the self-supervised learning strategy.

\begin{figure}[t]
  \centering
  \includegraphics[width=0.99\linewidth]{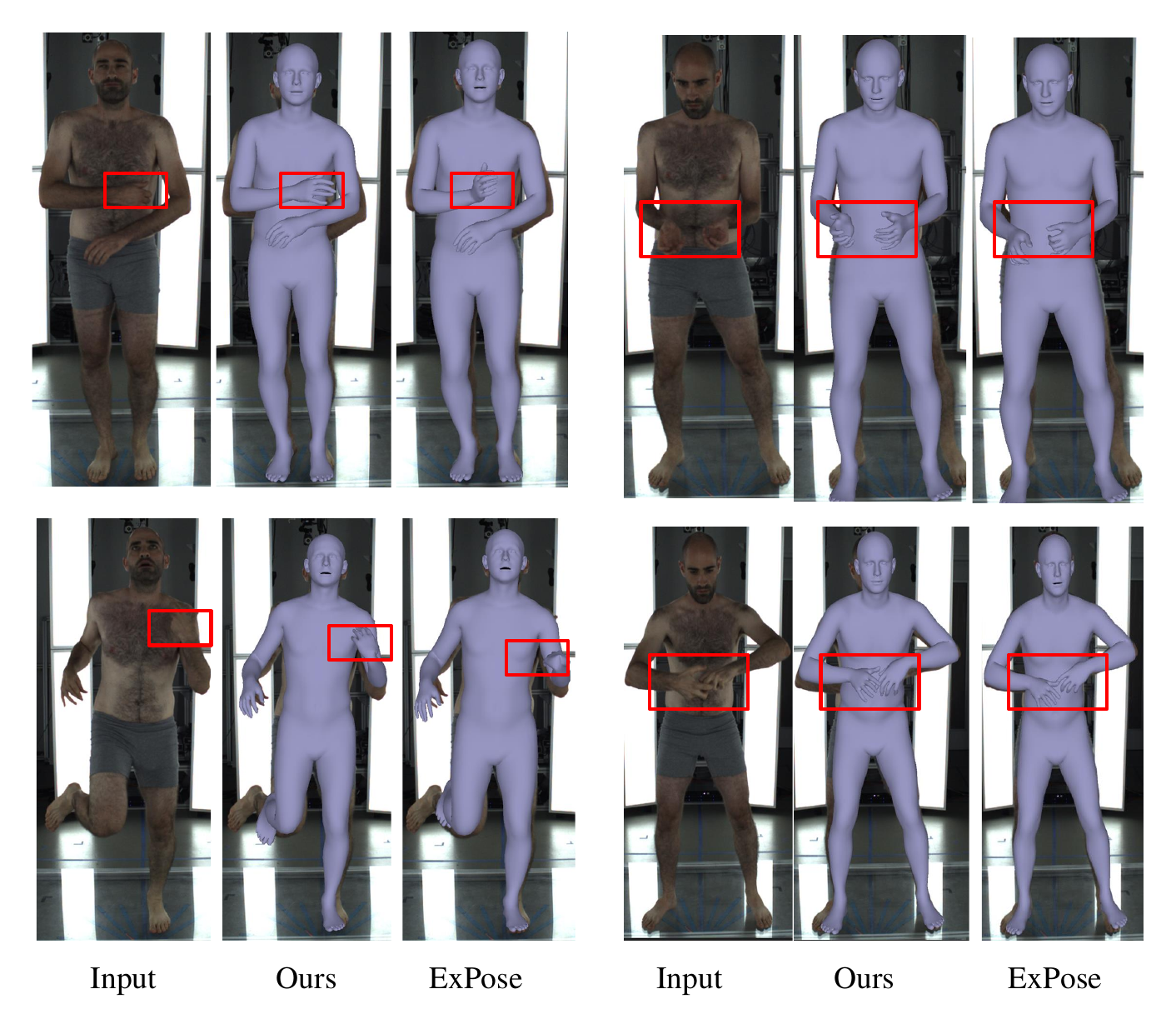}
  \vspace{-0.1in}
  \caption{Visualization results on EFH dataset.}
  \label{fig:visualization}
  \vspace{-0.1in}
\end{figure}

\begin{table}
\centering
\caption{Body reconstruction evaluation and comparison  on 3DPW dataset.}
\vspace{-0.1in}
\begin{tabular}{c|cc}
\hline
   Methods & Procrustes MPJPE (mm)  & Pelvis MPJPE \\
\hline
    ExPose &       60.7 &       93.4 \\

    FrankMocap &         60.0 &       94.3 \\

     PIXIE &       61.3 &          - \\
     
     SPIN &       59.2 &       96.9 \\

Ours+ExPose&         \textbf{59.0} &   \textbf{90.2} \\
\hline
\end{tabular} 
\label{tab:3dpw} 
\end{table}

\begin{table}

\centering
\caption{Face reconstruction evaluation and comparison  on NoW dataset.}
\vspace{-0.1in}
\begin{tabular}{c|ccc}
\hline
   Methods & PA-P2S Median(mm) &       mean &        std \\
\hline
    ExPose &       1.38 &       1.74 &       1.47 \\

     PIXLE &       1.18 &       1.49 &       1.25 \\

Deep 3D Face &       1.23 &       1.54 &       1.29 \\

  3DDFA &       1.23 &       1.57 &       1.39 \\

     PRNet &        1.5 &       1.98 &       1.88 \\

      DECA &  {\bf 1.10} & {\bf 1.38} & {\bf 1.18} \\

    MGCNet &       1.31 &       1.87 &       2.63 \\

RingNet&       1.21 &       1.54 &       1.31 \\

Ours+ExPose &       1.18 &        1.50 &       1.27 \\
\hline
\end{tabular}  
\label{tab:now}
\vspace{-0.2in}
\end{table}

\begin{table*}
\centering
\caption{Hand reconstruction evaluation and comparison  on Freihands dataset.}
\vspace{-0.1in}
\begin{tabular}{c|cccc}
\hline
   Methods & PA-MPJPE(mm) & PA-V2V(mm) &      F@5mm &     F@15mm \\
\hline
ExPose &       12.2 &       11.8 &       0.48 &       0.92 \\

  MANO CNN &         11.0 &       10.9 &       0.52 &       0.93 \\

     PIXIE &       12.9 &       12.1 &       0.47 &       0.92 \\

Ours+ExPose &       12.6 &       12.2 &       0.46 &       0.92 \\
\hline
\end{tabular}  
\label{tab:Freihands}

\end{table*}

\begin{table*}[t]
\centering
\caption{Ablation study on EFH dataset}
\vspace{-0.1in}
\begin{tabular}{c|ccccc}
\hline
   Methods & V2V/Procrustes & PA-V2V Body & PA-V2V Left Hand & PA-V2V Right Hand & PA-V2V Face \\
\hline
    ExPose &       55.1 &       52.9 &       13.1 &       12.6 &        5.7 \\

+ Psudeo ground-truth &       53.8 &       52.5 &       12.8 &       12.7 &        5.3 \\

+ Psudeo GT Selection &       52.9 &       52.5 &       12.7 &       12.9 &        5.3 \\

     + EMA(Ours+ExPose)&       \textbf{50.6}	&\textbf{51.6}&	\textbf{12.4}&	\textbf{11.7}&	\textbf{5.1}\\
\hline
\end{tabular}  
\label{tab:ablation}
\vspace{-0.15in}
\end{table*}

\subsection{Full-body Evaluation}

We first evaluate the performance of our method on full-body reconstruction. The experiments are conducted on EFH dataset. We follow \cite{pavlakos2019expressive} to use Vertex-To-Vertex/Procrustes, PA-V2V Body,	PA-V2V Left Hand,	PA-V2V Right Hand, and	PA-V2V Face metrics to evaluate the performance of the models.

\noindent \textbf{Comparison with the baseline:} The comparison results between our proposed methods and the baseline model ExPose are presented in Table \ref{tab:efh}. Several conclusions can be drawn from these results: 1) Our FuPRE significantly improves the performance of ExPose in terms of body, head, and hands reconstruction. Specifically, the Vertex-To-Vertex/Procrustes metric improves from 55.1 to 50.6, demonstrating a substantial advancement in the full-body reconstruction task. 2) The substantial improvement mainly originates from the superior reconstruction of body parts. This suggests that the body part reconstruction benefits most from our FuPRE. This is plausible given that human body parts are the most visible in training images, and thus, the pseudo labels would be the most precise. 3) Both variants of our method, namely FuPRE+ExPose and FuPRE+ResNet, showcase outstanding performance. Particularly, it's worth noting that our FuPRE+ResNet, a one-stage method, achieves superior results compared to the two-stage ExPose, even without the need of specifically cropping out face, hand and body parts for separate processing. This validates the effectiveness of our FuPRE training framework and demonstrates its superiority over traditional two-stage methods.

\noindent \textbf{Comparison with SOTA:} We further compare our proposed methods with the state-of-the-art (SOTA) method, mainly with PIXLE, as illustrated in Table \ref{tab:efh}. 1) For body part reconstruction, both our two-stage method (FuPRE+ExPose) and one-stage method (FuPRE+ResNet) surpass PIXLE. Specifically, our two-stage method achieves a Vertex-To-Vertex/Procrustes metric of 50.6, and our one-stage method reaches 54.8, both outperforming the score of 55.0 attained by PIXLE. This demonstrates the superior performance of our methods in body part reconstruction. 2) For face and hand parts reconstruction, our methods are comparable to PIXLE. Our two-stage method obtains a score of 5.1 on face part and 12.4 on hand parts, while the one-stage method achieves 5.6 on face part and 12.4 on hand parts. Compared to PIXLE's scores of 4.6 on face part and 11.0 on hand parts, our methods demonstrate competitive performance. These results confirm the robustness and effectiveness of our proposed methods, both in two-stage and one-stage scenarios. While we outperform PIXLE in body part reconstruction, we maintain competitive performance for face and hand parts. This demonstrates the balance our methods have achieved between performance and generalization across different body parts.

\noindent \textbf{Qualitative comparison:} Fig. \ref{fig:visualization} presents a visual comparison between our two-stage method and ExPose. A significant observation from this comparison is the pronounced accuracy of our method in predicting wrist poses. This can be attributed to the extensive use of pseudo labels in our network training process, enabling our models to learn and represent the pose distribution more effectively compared to ExPose. Moreover, our method exhibits superior performance in head reconstruction. For instance, in the bottom-right images of Fig. \ref{fig:visualization}, the subject's mouth is accurately depicted as closed by our method, whereas ExPose erroneously predicts it as open. This highlights the enhanced precision of our method in interpreting and reconstructing complex facial expressions. In essence, our method benefits substantially from the "learning from experts" pipeline, leading to superior performance in full-body reconstruction. This not only validates the effectiveness of our approach but also underscores the potential of utilizing pseudo labels to improve the learning process and the final reconstruction results.

\noindent \textbf{Performance Improvements through Incremental Data Addition:} The underlying premise of FuRPE is the effective utilization of pseudo labels and expert-derived features, generated by part-experts, to facilitate model training. This presents an interesting proposition: as the volume of data and pseudo ground-truths increases, the model's performance should correspondingly improve. To empirically validate this hypothesis, we incrementally augment the training set with additional data and observe the ensuing changes in performance. The results, as shown in Table \ref{tab:data}, lend credence to our hypothesis, unambiguously demonstrating that an increase in the volume of training data corresponds to enhanced model performance. This can be attributed to the model's ability to learn more generalized features and make increasingly accurate pose and shape predictions with a larger dataset.

 


\subsection{Part-only Evaluation}
\noindent \textbf{Body Part Evaluation:}  We assess the body-part reconstruction performance on the 3DPW dataset using Procrustes MPJPE and Pelvis MPJPE metrics, as shown in Table \ref{tab:3dpw}. Our method not only outperforms state-of-the-art full-body models but also excels over the body-centric SPIN method. This substantiates our method's superior body reconstruction capabilities, largely fostered by the use of pseudo labels and expert-derived features, thereby significantly improving upon the baseline.


\noindent \textbf{Head Part Evaluation:} We scrutinize the head-part reconstruction performance on the NoW dataset, utilizing the PA-P2S Median(mm), mean, and standard deviation metrics. The empirical results, presented in Table \ref{tab:now}, reveal our method's superiority over all full-body reconstruction techniques, whilst achieving performance on par with specialized head-only reconstruction methodologies. Notably, our results closely align with the current state-of-the-art DECA method. This empirical evidence underscores the promising nature of our head-reconstruction results.

\noindent \textbf{Hand Part Evaluation:} Table \ref{tab:Freihands} presents the hand-part reconstruction performance using PA-MPJPE(mm), PA-V2V(mm), F@5mm, and F@15mm metrics. All methods, including ours, exhibit similar performance, likely due to the low-resolution of training images limiting the model's learning capacity for hand information. Future efforts should focus on incorporating high-resolution images into the training and testing sets to enhance hand reconstruction performance and benchmarking.

\subsection{Ablation Study}

In Table \ref{tab:ablation}, we dissect the performance increments attributed to each key component added into our baseline network, ExPose. The implementation of the pseudo ground-truth generation module (including both pseduo labels and expert-derived features) reduces the overall V2V/Procrustes from 55.1 to 53.8, indicating the significant role of high-quality training data. Further inclusion of the pseudo ground-truth selection module refines the performance to 52.9, verifying its effectiveness in providing superior pseudo ground-truth. 

The introduction of the MEA training scheme, denoted as "+ EMA (Ours+ExPose)" in the table, brings about the most dramatic performance enhancement, reducing the V2V/Procrustes to 50.6, the lowest among all configurations. This performance leap underscores the MEA scheme's capability in imposing strong constraints through a self-supervised learning framework.

Comparing "Ours+ExPose" with other variants, it's evident that the integration of all components results in the most pronounced performance improvement across all metrics, including PA-V2V for Body, Left Hand, Right Hand, and Face. This observation affirms the collective indispensability of all components, underscoring the superiority of our comprehensive approach.

\section{Conclusion}
In this paper, we present FuRPE, a framework that tackles the scarcity of annotated data in full-body reconstruction. By utilizing part-experts and a pseudo ground-truth selection scheme, FuRPE generates high-quality pseudo labels, enhancing the learning process. Our novel training strategies further optimize the model's robustness. FuRPE significantly surpasses existing methods on multiple benchmarks, demonstrating its potential to redefine the state-of-the-art.

\bibliography{aaai24}
\end{document}